\documentclass[10pt,twocolumn,letterpaper]{article}

\usepackage{wacv}
\usepackage{times}
\usepackage{epsfig}
\usepackage{graphicx}
\usepackage{amsmath}
\usepackage{amssymb}
\usepackage[square,numbers,sort&compress]{natbib}

\usepackage{subfigure}
\usepackage{multirow}
\usepackage[inline]{enumitem}

\wacvfinalcopy 
\def\assignedStartPage{1}

\usepackage[breaklinks=true,bookmarks=false]{hyperref}

\begin{document}
\newcommand{\htmltag}[1]{\textless#1\textgreater}

%%%%%%%%% TITLE
\title{CoVA: Context-aware Visual Attention for Webpage Information Extraction}

\author{
    Anurendra Kumar$^*$\\
    {\tt\small ak32@illinois.edu}
    % For a paper whose authors are all at the same institution,
    % omit the following lines up until the closing ``}''.
    % Additional authors and addresses can be added with ``\and'',
    % just like the second author.
    % To save space, use either the email address or home page, not both
    \and
    Keval Morabia$^*$\\
    {\tt\small morabia2@illinois.edu}
    \and
    Jingjin Wang\\
    {\tt\small jingjin9@illinois.edu}
    \and
    Kevin Chen-Chuan Chang\\
    {\tt\small kcchang@illinois.edu}
    \and
    Alexander Schwing\\
    {\tt\small aschwing@illinois.edu}
    \and
    University of Illinois at Urbana-Champaign
}

\maketitle
\def\thefootnote{*}\footnotetext{These authors contributed equally to this work}\def\thefootnote{\arabic{footnote}}
%\thispagestyle{empty}

%%%%%%%%% ABSTRACT
\begin{abstract}
Webpage information extraction (WIE) is an important step to create knowledge bases. For this, classical WIE methods leverage the Document Object Model (DOM) tree of a website. However, use of the DOM tree poses significant challenges as context and appearance are encoded in an abstract manner. To address this challenge we propose to reformulate WIE as a context-aware Webpage Object Detection task. Specifically, we develop a Context-aware Visual Attention-based  (CoVA) detection pipeline which combines appearance features with  syntactical structure from the DOM tree. To study the approach we collect a new large-scale dataset\footnote{CoVA dataset and code are available at \href{https://github.com/kevalmorabia97/CoVA-Web-Object-Detection}{github.com/kevalmorabia97/CoVA-Web-Object-Detection}} of e-commerce websites for which we manually annotate every web element with four labels: product price, product title, product image and background. On this dataset we show that the proposed CoVA approach is a new challenging baseline which improves upon prior state-of-the-art methods.
\end{abstract}

%%%%%%%%%%%%%%%%%%%%%%%%%%%%%%%%%%%%%%%%%%%%%%-------------------------------------------------------------------------%%%%%%%%%%%%%%%%%%%%%%%%%%%%%%%%%%%%%%%%%%%%%%%%%%%%%%%%
\section{Introduction}

\begin{figure*}[t]
\centering
\hfill
\subfigure[]{\includegraphics[height=0.3\linewidth]{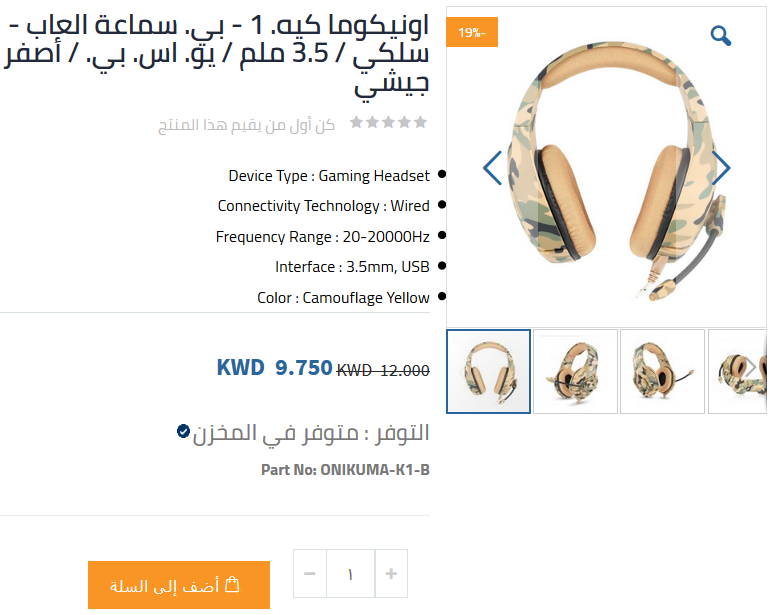}}
\hfill
\subfigure[]{\includegraphics[height=0.3\linewidth]{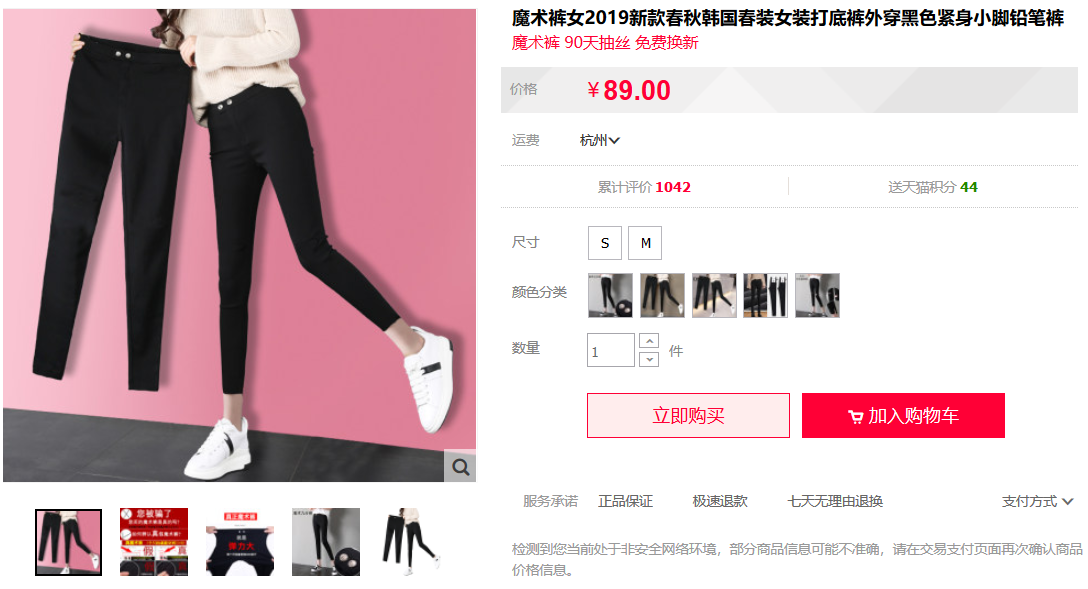}}
\caption{A person can detect the web element for product price, title, and image, even without knowing (a) Arabic or (b) Chinese}
\label{fig:chin_arabic}
\end{figure*}

Webpage information extraction (WIE) is an important step when creating a large-scale knowledge base \cite{chang2006survey, azir2017wrapper} which has many downstream applications such as  knowledge-aware question answering \cite{lin2019kagnet} and recommendation systems \cite{ma2019jointly, lin2020freedom}.

% {is there a good and well known example for WIE?}

Classical methods for WIE, like Wrapper Induction~\cite{soderland1999learning,muslea1998stalker,chang2001iepad}, rely on the publicly available source code of websites. The code is commonly parsed into a document object model (DOM) tree. The DOM tree is a programming language independent tree  representation of any website, which contains all its elements. It can be obtained using various libraries like Puppeteer\footnote{\href{https://pptr.dev}{https://pptr.dev}}. These elements contain information about their location in the rendered webpage, styling like font size, etc., and text if it is a leaf node. Various developer tools have been developed to inspect the DOM tree and modify it manually. 

However, using only the DOM tree for WIE is increasingly challenging for a variety of reasons: 1) Webpages are programmed to be aesthetically pleasing; 2) Oftentimes content and style is separated in website code and hence the DOM tree; 3) The same visual result can be obtained in a plethora of ways; 4) Branding banners and advertisements are interspersed with information of interest. %5) most importantly, website
%**-- 
For instance, the WIE,  WHISK \cite{soderland1999learning} which learns  extraction rules for number of bedrooms in a hotel  fails if  the style, content or tag   of  web element containing "bedroom" is changed. %---**
%{give an example of how an existing WIE which uses the DOM tree fails} 

For this reason, more recently, WIE applied optical character recognition (OCR) on rendered websites followed by word embedding-based natural language extraction \cite{staar2018corpus}. However, as mentioned before, recent webpages are highly enriched with visual content, and classical word embeddings don't capture this contextual information \cite{vishwanath2018deep}. For instance, text in advertising banners may be interpreted as  valuable information.  For this reason, a simple OCR detection followed by natural language processing techniques is a suboptimal WIE~\cite{vishwanath2018deep}. 

In response to these challenges we develop WIE based on a visual representation of a web element and its context. This permits to address the aforementioned four challenges. Moreover, visual features  are independent of the programming language (\eg, HTML for webpages, Dart for Android or iOS apps) and partially also the website language (\eg, Arabic, Chinese, English). Intuitively, we aim to mimic the ability of humans to detect the location of target elements like product price, product title and product image on a webpage in a foreign language like the one shown in Fig.~\ref{fig:chin_arabic}.

% \begin{figure*}[t]
% \centering
% \includegraphics[width=0.7\linewidth,height=3cm]{122_dom_squashed.pdf}
% \caption{DOM tree for webpage in Fig.~\ref{fig:multiple_prices}. Some nodes in the DOM tree are squashed if they only contain a single child for visualization}
% \label{fig:dom}
% \end{figure*}

% \begin{figure*}[t]
%     \centering
%     \includegraphics[trim=200 150 200 230,clip,width=\linewidth]{multiple_price_edge.png}
%     \caption{Example webpage showing multiple possible prices (red), but relatively fewer possible title (green) or image (purple)}
%     \label{fig:multiple_prices}
% \end{figure*}

\begin{figure*}[t]
    \centering
    \includegraphics[width=0.65\linewidth]{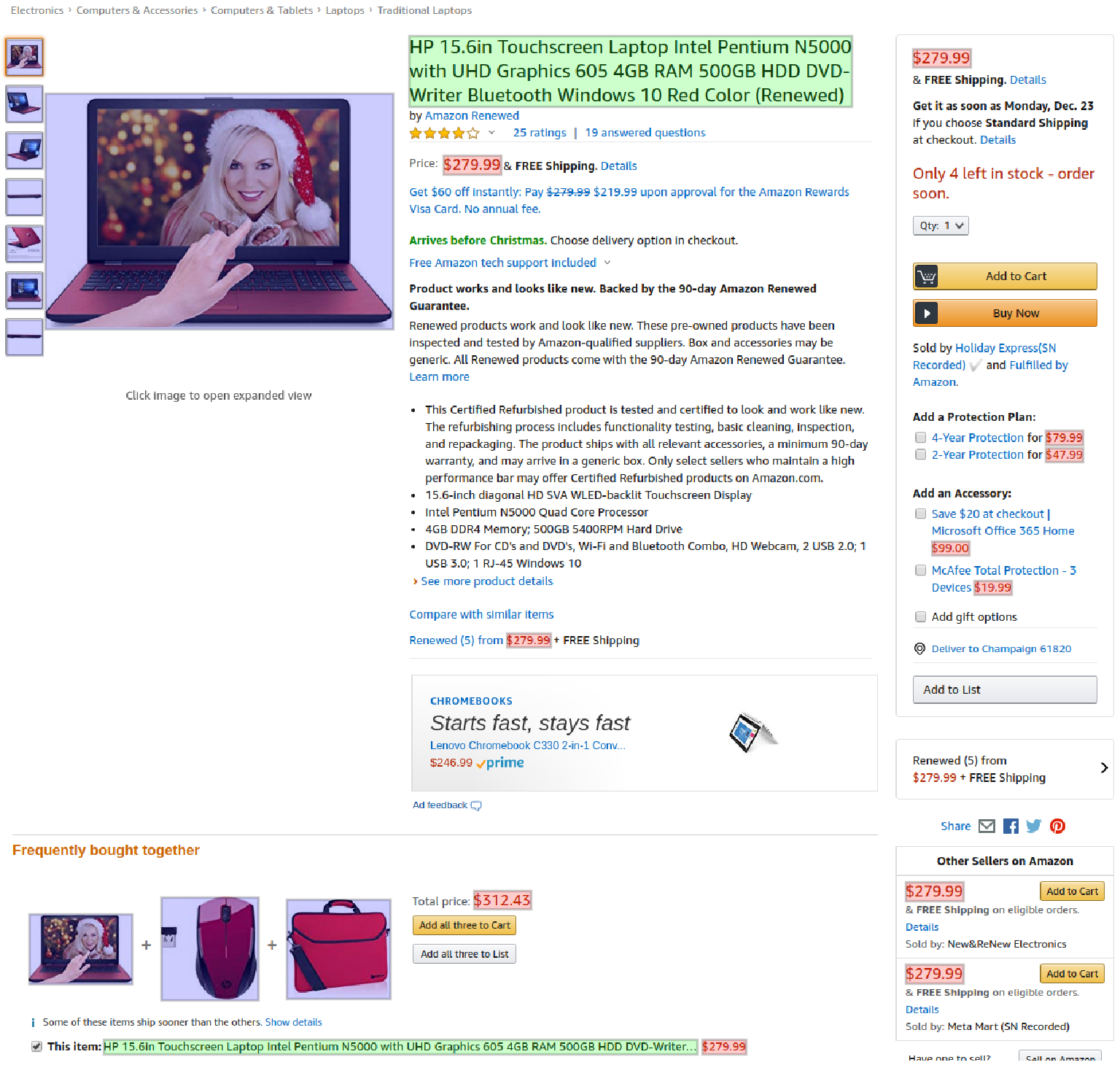}
    \caption{Example webpage showing multiple possible prices (red), but relatively fewer possible title (green) or image (purple)}
    \label{fig:multiple_prices}
\end{figure*}

For this, we develop a context-aware Webpage Object Detection (WOD), which we refer to as Context-aware Visual Attention-based detection (CoVA),  where entities like prices are objects. Somewhat differently from  an object in natural images which can be detected largely based on its appearance, objects on a webpage are strongly defined by contextual information. \Eg, a cat's appearance is largely independent of its nearby objects, whereas a product price is a highly ambiguous object (Fig.~\ref{fig:multiple_prices}). It refers to the price  of a product only when it is contextually related to a product title and a product image. The developed WOD uses a graph attention \cite{lin2017structured} based architecture, which leverages the underlying syntactic DOM tree \cite{zhou2021simplified}   to focus on important context \cite{zhu20052d} while classifying an element on a webpage. 

% {provide a one paragraph (5 line) summary of your experimental section}
To facilitate this task we create a dataset of $7.7$k product webpage screenshots along with DOM information spanning  $408$ different websites (domains). We create a cross-domain split so as to train on some domains (\eg, Amazon, Etsy), and evaluate on others (\eg, eBay). We compare the results of CoVA  with existing and newly created baselines that take visual features into account. For this we use accuracy of each class as the evaluation metric and  show that CoVA leads to substantial improvements while yielding interpretable contextual representations (see Sec.~\ref{fig:attn_viz}).

\noindent{}In summary, we make the following contributions: 
\begin{enumerate}[noitemsep,topsep=0pt,parsep=0pt,partopsep=0pt]
    \item We formulate WIE as a context-aware WOD problem.
    \item We develop a Context-aware Visual Attention-based (CoVA) detection  pipeline, which is  end-to-end trainable and exploits syntactic structure from the DOM tree along with screenshot images. CoVA uses a variant of Fast R-CNN \cite{girshick2015fast} to obtain a visual representation and graph attention \cite{graph_att}  for contextual learning on a graph constructed from the DOM tree. CoVA improves recent state-of-the-art %and newly created 
    baselines by a significant margin. 
    \item We create the largest public dataset of $7.7$k product webpage screenshots from $408$ online retailers for Object Detection from product webpages. Our dataset is $\sim 10\times$ larger than existing datasets.
    \item  We show the interpretability of CoVA using  attention visualizations (Sec.~\ref{subsec:attn_viz})
    %\item We claim and validate that visual features (without textual content) along with DOM information are sufficient for many tasks while allowing cross-domain and cross-language generalizability. CoVA trained on English webpages perform well on Chinese Webpages (Sec. \ref{sec:cross_lingual}).
\end{enumerate}

%%%%%%%%%%%%%%%%%%%%%%%%%%%%%%%%%%%%%%%%%%%%%%-------------------------------------------------------------------------%%%%%%%%%%%%%%%%%%%%%%%%%%%%%%%%%%%%%%%%%%%%%%%%%%%%%%%%

\section{Related Work}
\label{sec:related_work}
Webpage information extraction (WIE) has been mainly addressed with 
Wrapper Induction (WI). WI aims to learn a set of extraction rules from HTML code or text, using manually labeled examples and counter-examples \cite{soderland1999learning,muslea1998stalker,chang2001iepad}.  These often require human intervention which is time-consuming, error-prone  \cite{vadrevu2005automated}, and does not generalize to new templates.

Supervised learning, which treats WIE  as a classification task has also garnered significant attention. Structural and semantic features \cite{ibrahim2008automatic,gibson2007adaptive,xue2007web} are obtained for each part of a webpage to predict  categories like title, author, etc. \citet{wu2015automatic} cast  WIE as a HTML node selection problem using features such as positions, areas, fonts, text, tags, and links. \citet{joshi2009web} develop a semantic similarity between blocks of webpages using textual and DOM features to extract the key article on a webpage. \citet{rastogi2020information} extract visual cues which are followed by learning the relationship between elements. This information is utilized to allocate a document into predefined templates for which rules for target detection are learned using training data. \citet{lin2020freedom} proposes a neural network to learn representation of a DOM node by combining text and markup information. \cite{hwang2020spatial} develops a transformer architecture to learn spatial dependency between DOM nodes. Unlike these work which depends on text information, we aim to learn representation of a DOM node using only visual cues.

Visual features have been extensively employed to generate visual wrappers for pattern extraction. Mostly, these utilize hand-crafted visual features from a webpage, \eg,  area size, font size, and type \cite{cai2003extracting}. \citet{cai2003vips} develop a visual block tree of a webpage using visual and layout features along with the DOM tree information. Subsequent works use this tree for tasks like webpage segmentation, visual wrapper generation, and web record extraction \cite{cai2004block,liu2003mining,simon2005viper,burget2009web}.  \citet{zheng2007template} develop a supervised learning framework using predefined visual features to extract content from news webpages. \citet{gogar2016deep} aim to develop domain-specific wrappers which  generalize across unseen templates and don't need manual intervention. They develop a unified model that encodes visual features, textual features, and positional features using a single Convolutional Neural Net (CNN). 

Object detection (OD) which aims to detect and classify all objects, has been extensively studied for natural images. Deep learning-based methods such as YOLO \cite{redmon2018yolov3}, R-CNN variants \cite{girshick2014rich,girshick2015fast,ren2015faster,he2017mask}, SSD \cite{liu2016ssd}, etc.\  yielded state-of-the-art results in OD. OD methods that can capture contextual information are of particular interest here.  %\cite{mottaghi2014role} studies the significance of OD and segmentation tasks by creating a dataset which labels each pixel by its parts and develops a deformable parts model to capture context. 
\citet{divvala2009empirical} learn contextual information for presence, size and location of other objects. \citet{perko2007context} learn a context confidence score for each class by estimating the importance of each pixel. \citet{murphy2006object} learn local and global context by object presence and localization and use a product of experts model \cite{hinton2002training}  to combine them. \citet{torralba2001statistical} learn global context information in terms of the spatial layout of spectral components. \citet{cinbis2012contextual} learn the set of descriptors for other objects in the scene and learn intra-class and inter-class spatial relations.

Graph Convolutional Networks (GCN) \cite{kipf2016semi} was proposed to learn a node representation while taking neighbors of a node into account. Using it, \citet{graph_multimodal} represent a visually rich document as a complete graph of text content obtained by passing OCR \cite{mithe2013optical}. They employ GCN to learn node representations for each web element. 
% text embedding - BILLSTM CRF, and which is followed by BiLSM CRF on sequence of represntation for classication.

Recently, attention mechanisms have also shown remarkable ability in capturing contextual information  \cite{bahdanau2014neural}. \citet{vaswani2017attention} propose a transformer architecture  for language modeling. Word vectors learned on BERT \cite{devlin2018bert}, which use self-attention, have yielded state-of-the-art results on $11$ NLP tasks. \citet{luo2018attention} use attention over a BiLSTM-CRF layer for Named Entity Recognition (NER) on biomedical data. Separately, attention  has been used for contextual learning in OD \cite{li2013web,hsieh2019one,morabia2020attention} and image captioning \cite{you2016image}. Attention mechanisms have also  been employed over graphs to learn an optimal representation of nodes while taking graph structure into account \cite{graph_att}.

Moreover, attention permits to  interpret result, which is often desired in many applications. We show our visualizations depicting this advantage below (Sec.~\ref{subsec:attn_viz}).
%------------------------------------------------------------------

\begin{figure*}[t]
\vspace{-0.3cm}
    \centering
    \includegraphics[trim=57 70 97 52,clip,width=\linewidth]{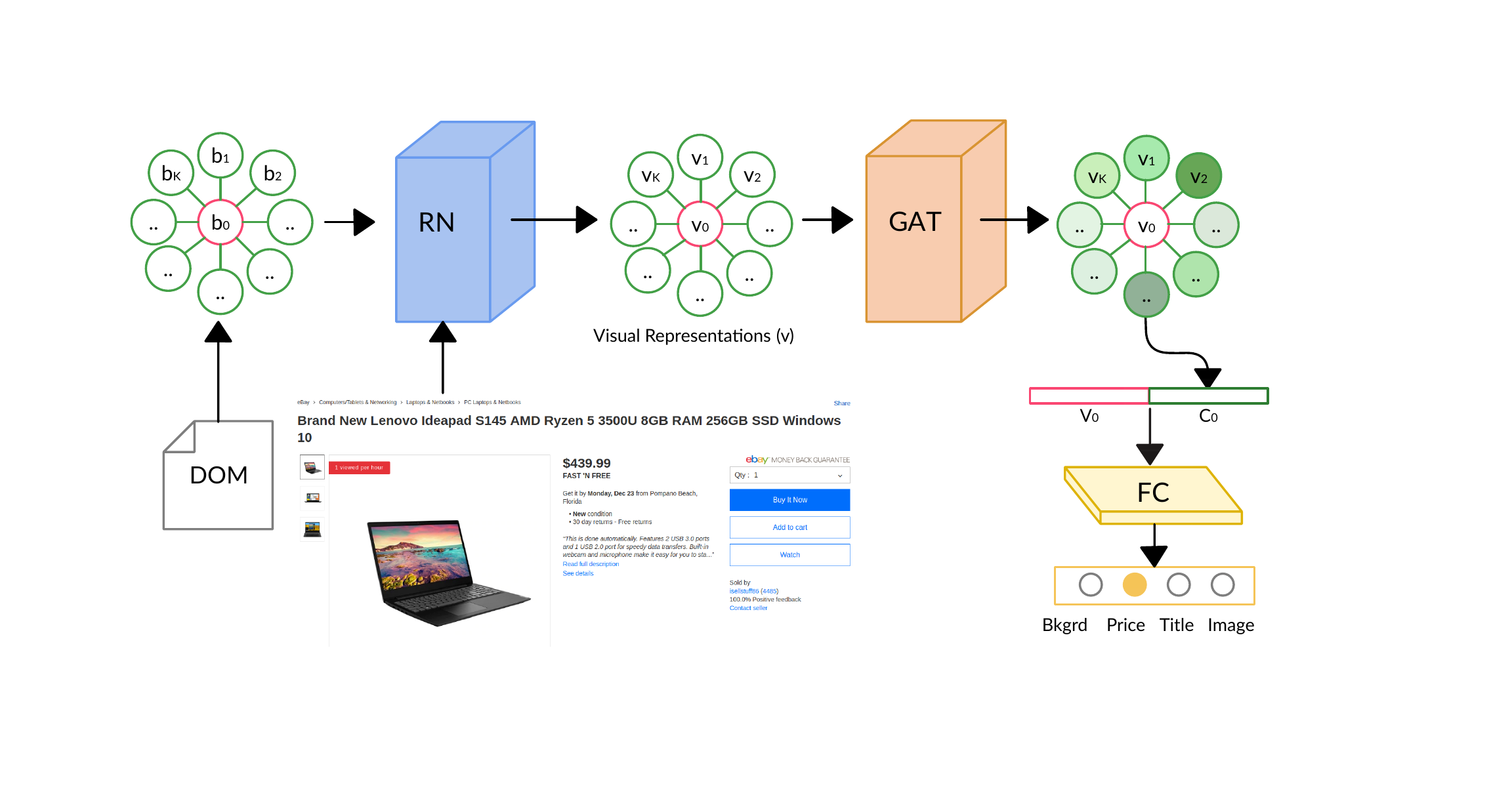}
    \caption{CoVA end-to-end training pipeline (for a single web element). CoVA takes a webpage screenshot and list of bounding boxes  along with $K$ neighbors for each web element (obtained from DOM). RN learns visual representation ($v_0$) while GAT learns contextual representation ($c_0$) from its neighbor's visual representations. $v_0$ and $c_0$ are concatenated and passed through FC layer.}
    \label{fig:ModelArchitecture}
\vspace{-0.3cm}
\end{figure*}

% Training architecture for CoVA. CoVA takes webpage screenshot, and bounding boxes of a web element to be classified and its $ K $ syntactically closer context elements (obtained from DOM tree) as input. Representation Network (RN) learns a visual representation of all these elements using a CNN followed by RoI pooling, and a Positional encoder. Graph Attention Network (GAT) learns relevance score for each context element using Self-Attention to compute a contextual representation for the web element to be classified. Finally, visual and contextual representations are passed through a feed forward network to output class probabilities

%%%%%%%%%%%%%%%%%%%%%%%%%%%%%%%%%%%%%%%%%%%%%%-------------------------------------------------------------------------%%%%%%%%%%%%%%%%%%%%%%%%%%%%%%%%%%%%%%%%%%%%%%%%%%%%%%%%

\section{Problem formulation}
\label{sec:problem_formulation}

% \subsection{Problem Formulation}
The DOM tree captures the syntactical  structure of a webpage %which captures syntactical structure 
similar to a parse tree of a natural language. 
Our goal is to extract semantic information exploiting this syntactic structure. We view a leaf web element as a \textit{word} and the webpage as a document with the DOM tree as its underlying parse tree. %Ideally, we would like to develop an algorithm that predicts the probability of a web element being target given its own representation and the DOM tree. 
Formally, we  represent a webpage $W$ as the set $W=\{v_1,v_2,\dots,v_i,\dots,v_N, D\}$ where $v_i$ denotes the visual representation of the $i$-th web element, $N$ denotes number of web elements, and $D$ refers to the DOM tree which contains the relations between the web elements.  Our goal is to learn  a parametric function $f_\theta(y_i|W,i)$ which extracts a visual representation $v_i$ of the $i$-th web element from website $W$ so as to  accurately predict label $y_i$ of the web element. In the following we consider four labels, \ie,  $y_i\in\{\text{product price}, \text{title}, \text{image}, \text{background}\}$. The parameters $\theta$ are obtained by minimizing the following supervised classification loss  
\[\theta^\ast =  \underset{\theta}{\mathrm{argmin}} \quad  
\underset{i, W {\sim} P_W}{\mathbb{E}} \left[\mathcal{L}(f_\theta(y_i|W,i),{y}^{*}_i)\right],\]
where $\mathbb{E}$ denotes an expectation, $y_i$ and $y^*_i$ denote the predicted and ground truth  labels and $P_W$ denotes a probability distribution over webpages.

%\subsection{WIE as a Context-aware Object Detection}
Information of a webpage is present in the leaves of the DOM tree, \ie, the web elements $i$. Web elements are an atomic entity which is characterized by a rectangular bounding box.  We can extract the target information $y_i$ from the DOM tree if we know the exact leaf bounding boxes of the  desired element. Therefore, we can view WIE as an object detection (OD) task where objects are  leaf elements. % and might contain the desired entity (target).  
However, identity $y_i$ of a web element is heavily dependent on its context, \eg, price, title, and image of a product are most likely to be in same or nearby sub-tree in comparison to unrelated web elements such as advertisements. Similarly, there can be multiple instances of price-like elements. However, the correct price would be contextually positioned with product title and image (Fig.~\ref{fig:multiple_prices}). Therefore, we formulate WIE as a context-aware object detection. 

We use the DOM tree to identify context for a web element.  We represent the syntactic closeness between web elements  through edges in the graph (discussed in next section). We then employ a graph attention mechanism \cite{graph_att} to attend to the most important contexts.

%-----------------------------------------------------##---------
%-----------------------------------------------------##---------
\section{Proposed End-to-End Pipeline -- CoVA}

In this section, we present our Context-Aware Visual Attention-based end-to-end pipeline for Webpage Object Detection (CoVA) which aims to learn function $f$ to predict labels  $y = [y_1,y_2,\dots,y_N]$ for a webpage.
%{remind people of function $f$ and its output}
The input to CoVA consists of \begin{enumerate*}
    \item a screenshot of a webpage,
    \item list of bounding boxes $[x,y,w,h]$ of the web elements, and
    \item neighborhood information for each element obtained from  DOM. %an adjacency list of each node (web element).
\end{enumerate*} 

As illustrated in Fig.~\ref{fig:ModelArchitecture} %shows our end-to-end architecture. T
this information is processed by the developed 
CoVA in four stages: \begin{enumerate*}
    \item the graph representation extraction for the webpage,
    \item the Representation Network (RN),
    \item the Graph Attention Network (GAT), and
    \item a fully connected (FC) layer.
\end{enumerate*} 
The graph representation extraction computes for every web element $i$ its set of neighboring web elements ${\cal N}_i$.
The RN consists of a Convolutional Neural Net (CNN) and a positional encoder aimed to learn a visual representation $v_i$ for each web element $i\in\{1, \dots, N\}$. The GAT combines the visual representation $v_i$ of the web element $i$ to be classified and those of its neighbors, \ie, $v_k$ $\forall k\in{\cal N}_i$ to compute the contextual representation $c_i$ for  web element $i$. Finally, the visual and contextual representations of the web element are concatenated and passed through the FC layer to obtain the classification output. We describe each of the components next.

\subsection{Webpage as a Graph}

As discussed earlier, the identity of a web element depends on its context. Therefore, we represent a webpage as a graph where nodes are leaf web elements and an edge indicates that the corresponding web elements are contextually relevant to each other. The graph representation of a webpage permits to learn a contextual representation of a web element by identifying important context, \eg, the currency symbol  near a price. %A naive way to create graph is by putting edge between every pair of nodes \cite{graph_multimodal}. An alternative way of creating a graph is to add edges to nearby nodes based on spatial distance. However, web elements vary greatly in shapes \& sizes, and two web elements might have low distance but they're contextually irrelevant since they lie in different DOM subtrees. 
%For this we use  a graph attention \cite{graph_att} and represent a webpage as a graph. 
An edge within the graph denotes the syntactic closeness in the DOM tree. Specifically,  we use the $K $ nearest leaf elements in the DOM tree as the neighbors ${\cal N}_i$ of a web element $i$.

\subsection{Representation Network (RN)}
\label{sec:rn}
The goal of the Representation Network (RN) is to learn a fixed size visual representation $v_i$ of any web element $i\in\{1, \dots, N\}$. This is important 
since  web elements have different sizes, aspect ratios, and content type (image or text). To achieve this the RN consists of a CNN operating on the screenshot of a webpage, followed by a Region of Interest (RoI) pooling layer \cite{girshick2015fast} and a positional encoder. Specifically, RoI pooling is performed to obtain a fixed size representation for all web elements. To capture the spatial layout, we learn a $P$ dimensional positional feature which is obtained by passing the bounding box features $[x, y, w, h, \frac{w}{h}]$ through a positional encoder implemented by a single layer neural net. Finally, we concatenate the flattened output of the RoI pooling with positional features to obtain the visual representation $v_i$. %and call it Visual Representation ($v_i$) for the $i$-th web element.

\subsection{Graph Attention Network (GAT)}
The goal of the graph attention network is to compute a contextual representation $c_i$ for each web element $i$ which takes information from neighboring web elements into account. 
However, out of multiple neighbors for a web element, only a few  are  informative,  \eg, a web element having a currency symbol  near a set of digits seems relevant. To identify the relational importance we use a Graph Attention Network (GAT) \cite{graph_att}. It takes the visual representations $v_i$ of a web element and its neighbors, and computes the contextual representation $c_i$. Formally, let $v = [v_1,v_2,\dots,v_N]$ represent the visual representations of web elements obtained from the RN. We transform each of the input features by learning projection matrices $W_1$ and $W_2$ applied at every node and its neighbors. We then employ self-attention \cite{lin2017structured} to compute the importance score,
\begin{equation}
    \alpha_{ij} = \frac{\exp(\text{LeakyReLU}(a^T[W_1 v_i || W_2 v_j]))}{\sum_{k \in \mathcal{N}_{i}} \exp(\text{LeakyReLU}(a^T[W_1 v_i || W_2 v_k]))},
\end{equation}
where $\cdot^T$ represents transposition, $||$ is the concatenation operation, $\mathcal{N}_i$ denotes the neighbors of web element $i$. The weights $\alpha_{ij}$ are non-negative attention scores for neighboring web elements of web element $i$. Finally, we obtain the contextual representation $c_i$ for a web element $i$ as a weighted combination of projected visual representations of its neighbors, \ie, via
\begin{equation}
% \begin{split}
{c_i} = \sum_{j \in \mathcal{N}_{i}} \alpha_{ij} W_2 v_j.
% \sum\limits_{k=1}^N \alpha_{k} = 1, \quad \alpha_{k} \geq 0
% \end{split}
\end{equation}

\subsection{Augmenting CoVA with extra features}
\label{sec:CoVA++}
In  scenarios where additional features (\eg, text content, HTML tag information, etc.) are available, CoVA can be easily extended to incorporate those. These features can  be concatenated with visual representations obtained from the RN without   modifying the model in any other way. We refer to this extended model as \textbf{CoVA++}. However, making the model dependent on these features might lead to constraints regarding the  programming language (HTML tags) or text language. In Sec.~\ref{sec:cross_lingual}, we show that CoVA trained on English webpages (without additional features) generalizes well to Chinese webpages. This result suggests that CoVA is able to learn  visual representations that are generalizable. %Hence, depending on the requirement, the variant of CoVA can be used.

%-----------------------------------------------------##---------
%-----------------------------------------------------##---------

\section{Dataset Generation}
%So far, there had been a scarcity of any large scale dataset on which machine learning can be performed for WIE. 
To the best of our knowledge there is no large-scale dataset for WIE with visual annotations for object detection. So far, the Structured  Web  Data  Extraction (SWDE) dataset \cite{hao2011one} is the only known large dataset that can be used for training deep neural networks for WIE \cite{lin2020freedom,lockard-etal-2019-openceres}. SWDE dataset contains webpage HTML codes  which is not sufficient to render it into a screenshot (since it contains links to old and non-existent URLs). Because of this we create a new large-scale labeled dataset for object detection on product webpage screenshots. We chose e-commerce websites since those have been a de-facto standard for WIE \cite{gogar2016deep,zhu20052d}. Our dataset generation consists of two steps: \begin{enumerate*} \item search the web with `shopping' keywords to aggregate diverse webpages and employ heuristics to automate labeling of product price, title, and image, \item manual correction of incorrect labels. \end{enumerate*} We discuss both steps next.

\noindent\textbf{Web scraping and coarse labeling.}
To scrape websites we use Google shopping\footnote{\href{https://shopping.google.com/}{shopping.google.com}} which aggregates links to multiple online retailers (domains) for the same product. These links are uploaded by the merchants of the respective domains. We do a keyword search for various categories, like 'electronics,' 'food,' 'cosmetics.' For each search result, we record the price and title from Google shopping. Then, we navigate through the links to specific product websites and save a $1280 \times 1280$ screenshot. To extract a bounding box for each web element, we store a pruned  DOM tree. Price and title candidates are labeled by comparing with the recorded values using heuristics. For product images, we always choose the  DOM element having the largest bounding box area among all the elements with an \htmltag{img} HTML tag, although this might not be true for many websites. We correct this issue in the next step.

\noindent\textbf{Label correction.}
The coarse labeling is only ${\sim}60\%$ accurate because \begin{enumerate*} \item price on   webpages keeps changing and might  differ from the Google shopping price, and \item many bounding boxes have the same content. \end{enumerate*} To correct for these mistakes, we manually inspected and correct labeling errors. In many cases, product price or product title is present multiple times in a webpage, so we made our best effort to choose the best one given its context. %This step is simply to identify the real ground truth in an unbiased manner. 
We obtain  \textbf{7,740 webpages} spanning  \textbf{408 domains}. Each of these webpages contains exactly one labeled price, title, and image. All other web elements are labeled as background. On average, there are ${\sim}90$ web elements on a webpage.

\noindent\textbf{Train-Val-Test split.}  
We create a cross-domain split which ensures that each of the train, val and test sets contains webpages from different domains. Specifically, we construct a $3:1:1$ split  based on the number of distinct domains. We observed that the top-5 domains (based on number of samples) were Amazon, EBay, Walmart, Etsy, and Target. So, we created 5 different splits for 5-Fold Cross Validation such that each of the major domains is present in one of the 5 splits for test data.

%%%%%%%%%%%%%%%%%%%%%%%%%%%%%%%%
%%%%%%%%%% MAIN TABLE %%%%%%%%%%
%%%%%%%%%%%%%%%%%%%%%%%%%%%%%%%%
\begin{table*}[ht]
\begin{center}
    \begin{tabular}[h]{| l | c | c | c | c |}
    \hline
    % \multicolumn{5}{| c |}{Cross Domain Accuracy}\\
    % \hline
    Method & No. of parameters & Price Accuracy & Title Accuracy & Image Accuracy \\
    \hline\hline
    Gogar \etal \cite{gogar2016deep} & 1.8m &  $ 78.1 \pm 17.2 $ &  $ 91.5 \pm 1.3 $ & $ 93.2 \pm 1.9 $ \\
    Random Forest using Heuristic features  & - &  $ 87.4 \pm 10.4 $ &  $ 93.5 \pm 5.3 $ & $ 97.2 \pm 3.8 $ \\
    Fast R-CNN* \cite{girshick2015fast}  & 0.5m &  $ 86.6 \pm 7.3 $ &  $ 93.7 \pm 2.2 $ & $ 97.0 \pm 3.6 $ \\
    % Fast R-CNN* + Averaged Context  & 1.6m &  $ 86.9 \pm 10.7 $ &  $ 94.7 \pm 1.7 $ & $ 97.6 \pm 2.9 $ \\
    Fast R-CNN* + GCN \cite{kipf2016semi}   & 1.4m &  $ 90.0 \pm 11.0 $ &  $ 95.4 \pm 1.5 $ & $ 98.2 \pm 2.8 $ \\
    Fast R-CNN* + Bi-LSTM \cite{schuster1997bidirectional} & 5.1m &  $92.9 \pm 4.6$ &  $94.0 \pm 2.1$ & $ 97.6 \pm 3.6 $ \\
    \hline
    \textbf{CoVA}  & 1.6m &  $ \mathbf{95.5} \pm \mathbf{3.8} $ &  $ \mathbf{95.7} \pm \mathbf{1.2} $ & $ \mathbf{98.8} \pm \mathbf{1.5} $ \\
    % \textbf{CoVA++}  & 2.6m &  $ \mathbf{96.1} \pm \mathbf{3.8} $ &  $ \mathbf{96.8} \pm \mathbf{1.9} $ & $ \mathbf{99.3} \pm \mathbf{0.5} $ \\
    \textbf{CoVA++}  & 1.7m &  $ \mathbf{96.1} \pm \mathbf{3.0} $ &  $ \mathbf{96.7} \pm \mathbf{2.2} $ & $ \mathbf{99.6} \pm \mathbf{0.3} $ \\
    \hline
    \end{tabular}
\end{center}
\vspace{-0.3cm}
\caption{Cross Domain Accuracy (mean $ \pm $ standard deviation) for 5-fold cross validation. }
\label{tab:main_comparison}
%\vspace{-0.3cm}
\end{table*}

%-----------------------------------------------------##---------
%-----------------------------------------------------##---------
\section{Experimental Setup \& Results}
\label{sec:exp}

In this section, we  present our experimental setup, evaluation metrics, comparison of results with the baselines, and attention visualizations of our model.

% \subsection{Baseline Methods} As discussed in Sec.~\ref{sec:related_work}, Wrapper Induction %methods are not scalable and 
% does not generalize to previously unseen web templates \cite{chang2006survey, lin2020freedom}. Hence we don't consider Wrapper Induction as a baseline. Instead, we 
\subsection{Baseline Methods} 
We compare the results  of our end-to-end pipeline CoVA with other existing  and newly created baselines summarized below. Our newly created baselines combine existing object detection and graph based models to identify the importance of visual features and contextual representations.

\noindent\textbf{\cite{gogar2016deep}:} This method identifies product price, title, and image from the visual and textual representation of the web elements. We use their publicly available code to train it on our dataset. 
    
\noindent\textbf{Random Forest on Heuristic features:} We train a Random Forest classifier with $100$ trees using various heuristic HTML tag-based, text, and bounding box features. HTML tags like \htmltag{H1}, \htmltag{P}, \htmltag{IMG}, etc.\ are one-hot encoded. Textual features include font size, number of words, and binary features like presence of currency symbols, text, and number. Bounding box features  $ [x, y, w, h, \frac{w}{h}] $ for the web element are also used.

\noindent\textbf{Fast R-CNN*:} We compare with Fast R-CNN \cite{girshick2015fast}  to quantify the importance of contextual representations in CoVA. We use the  DOM tree instead of selective search \cite{uijlings2013selective} for bounding box proposals. Since the proposals are exactly localized on the webpage, there is no need for bounding box regression. We also use positional features as described when discussing the representation network (Sec.~\ref{sec:rn}) for a fair comparison with CoVA. We will refer to this baseline as `Fast R-CNN*.'
	
\noindent\textbf{Fast R-CNN* + GCN \cite{kipf2016semi}:} We use graph convolution networks on our graph formulation where node features are the visual representations obtained from Fast R-CNN*.
	
\noindent\textbf{Fast R-CNN* + Bi-LSTM \cite{schuster1997bidirectional}}: We train a bidirectional LSTM on visual representations of web elements in preorder traversal  of the DOM tree. We use its output as the contextual representation and concatenate it with the visual representation of the web element obtained from Fast R-CNN*.

\subsection{Model Training, Inference and Evaluation} In each training epoch, we randomly sample  $90\%$ from background (neither of price, title, or image) web elements.  This  increases the diversity in training data by providing different contexts for webpages with exactly the same template. Overall, it introduces stochasticity and reduces over-fitting for contextual learning while decreasing the number of computations.   We use batch normalization \cite{ioffe2015batch} between consecutive layers which improves convergence and final performance. We train the model for a maximum of $50$ epochs with early stopping, after which we restore model parameters to the epoch corresponding to the best validation data result. We use the Adam optimizer for updating model parameters and minimize cross-entropy loss. During inference, the model detects one web element with highest probability for each class. Once the web element is identified, the corresponding text content can be extracted from the DOM tree or by using OCR for downstream tasks.

For CoVA++ we use as additional information the same heuristic features used to train the Random Forest classifier baseline. Unless specified otherwise, all results of CoVA and baselines  use the following hyperparameters where applicable: learning rate = $5\textrm{e-4}$, batch size = $5$ screenshot images, $K=24$ neighbor elements in the graph, RoI pool output size $(H \times W) = (3\times3)$, dropout = $0.2$, $P$ = $32$ dimensional positional features, output dimension for projection matrix $W_1, W_2$ is $384$, weight decay = $1\textrm{e-3}$. We use the first $5$ layers of a pre-trained ResNet18 \cite{he2016deep} in the representation network (RN), which yields a $64$ channel feature map. This significantly reduces the parameters in the RN from $12m$ to $0.2m$ and speeds up training at the same time. % yielded optimal results. 
The evaluation is performed using Cross-domain Accuracy for each class, \ie, the fraction of webpages of new domains with correct class. All the experiments are performed on Tesla V100-SXM2-16GB GPUs.

\begin{figure}[t]
    \centering
    \includegraphics[width=0.8\linewidth]{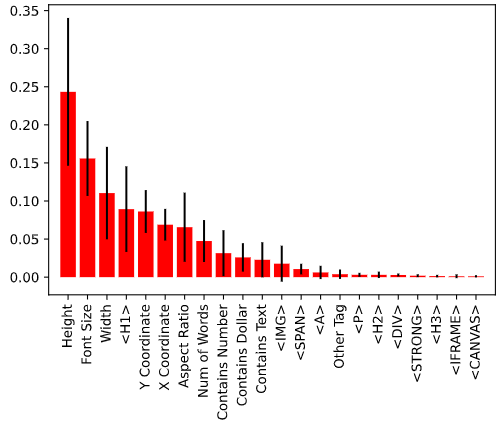}
    \caption{Gini impurity-based importance of features in RF}
    \label{fig:heuristic_feature_importances}
    \vspace{-0.3cm}
\end{figure}

\begin{figure*}
    \centering
    \subfigure[]{\includegraphics[width=0.6\linewidth]{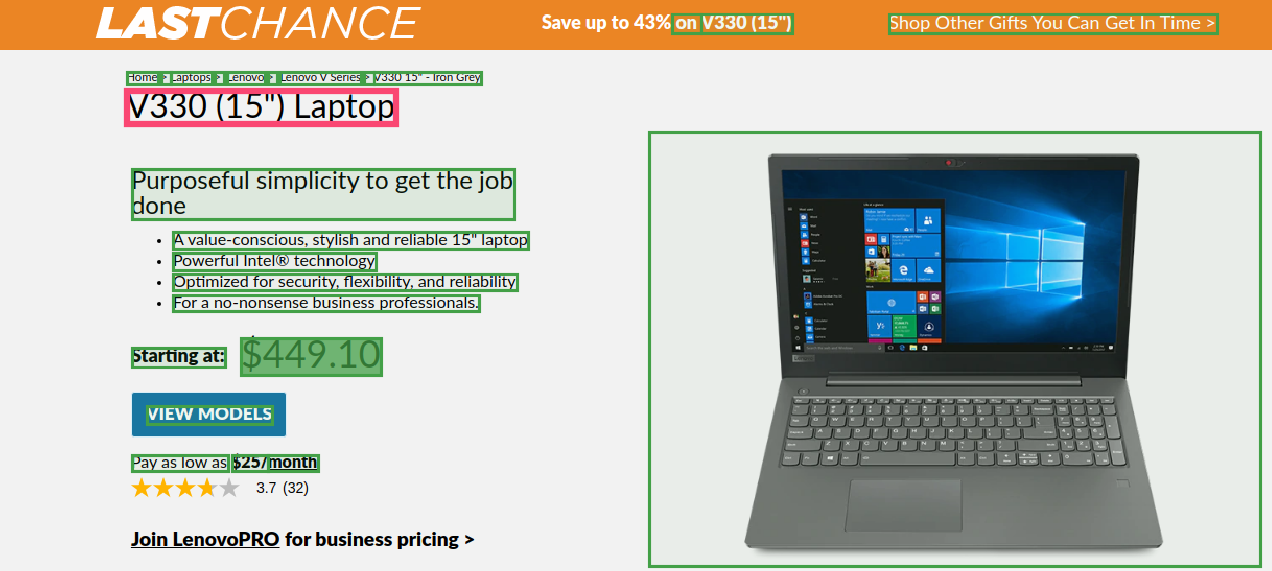}}
    \subfigure[]{\includegraphics[width=0.6\linewidth]{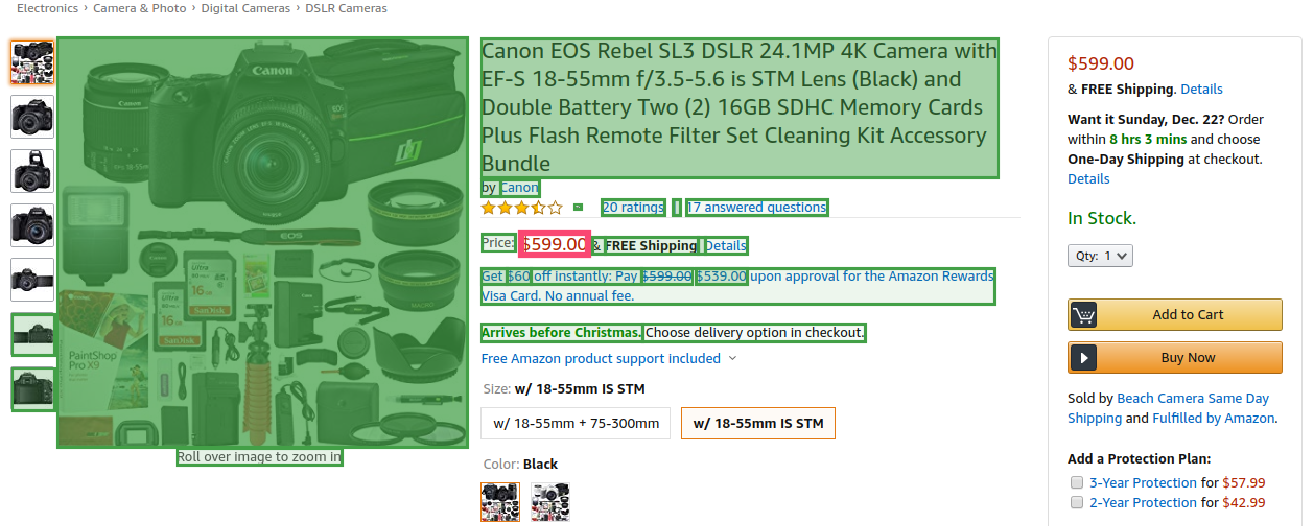}}
    \caption{Attention Visualizations where red border denotes  web element to be classified, and its contexts have green shade whose intensity denotes score. Price in (a) get much more score than other contexts. Title and image in (b) are scored higher than other contexts for price.}
    \label{fig:attn_viz}
\end{figure*}

%%%%%%%%%%%%%%%%%%%%%%%%%%%%%%5
\subsection{Results}

As shown in Table~\ref{tab:main_comparison}, our method outperforms all  baselines   by a considerable margin especially for price prediction. CoVA learns visual features which are significantly better than the heuristic feature baseline that uses predefined tag, textual and visual features.  Fig.~\ref{fig:heuristic_feature_importances} shows the importance of different heuristic based features in a webpage. We observe that a heuristic feature based method has similar performance to methods which don't use contextual features.  Moreover, CoVA++ which also uses heuristic features, doesn't lead to significant improvements. This shows that visual features learnt by CoVA are more general for tasks like price \& title detection. Context information is particularly important for price (in comparison to title and image) since it's highly ambiguous and occurs in  different locations with varying contexts (Fig.~\ref{fig:multiple_prices}). This is evident from the ${\sim}8.9\%$ improvement in price accuracy compared to the Fast R-CNN*. Unless stated otherwise, we will discuss results with respect to price accuracy. We observe that CoVA yields stable results across folds (${\sim}3.5\%$ reduction in standard deviation). This shows that CoVA learns features which are generalizable and which have less dependence on the training data. Using GCN with Fast R-CNN*  leads to unstable results with $11\%$ standard deviation while yielding a $3.4\%$ improvement over Fast R-CNN*.  Fast R-CNN* with Bi-LSTM is able to summarize the contextual features by yielding a ${\sim}6.3\%$ improvement in comparison to Fast RCNN*. CoVA outperforms Fast RCNN* with Bi-LSTM by ${\sim}2.6\%$ with much fewer number of parameters while also yielding interpretable results. We also obtained top-$3$ accuracy for CoVA, which are $98.6\%$, $99.4\%$, and $99.9\%$  for price, title and image respectively.

\subsection{Cross-lingual Evaluation of CoVA}
\label{sec:cross_lingual}
To validate our claim that visual features (without textual or HTML tag information) can capture cross-lingual information, we test our model on  webpages in a foreign language. In particular, we evaluated CoVA (trained on English product webpages) using $100$ Chinese product webpages spanning across $25$ unique domains. CoVA achieves $92\%$, $90\%$, and $99\%$ accuracy for product price, title, and image.  

\subsection{Attention Visualizations}
\label{subsec:attn_viz}

Table~\ref{tab:main_comparison} shows that attention significantly improves performance for all the three targets. As discussed earlier, only few of the contexts are important which are effectively learnt by our Graph Attention Network (GAT). We observed that on average, $ {\sim}20\% $ of context elements were activated (score above $ 0.05 $ threshold) by GAT. We also study a multihead attention instead of single head following \cite{vaswani2017attention}, which didn't yield significant improvements in our case.

Fig.~\ref{fig:attn_viz} shows visualizations of attention scores learnt by GAT. Fig.~\ref{fig:attn_viz}(a) shows an example where title and image have more weight than other contexts when learning a context representation for price. This shows that attention is able to focus on important web elements and discards others. Similarly, Fig.~\ref{fig:attn_viz}(b) shows that price has a much higher score than other contexts for learning contextual representation for title. We found that there are some cases where attention gives similar weights to all contexts.

%-----------------------------------------------------##---------
\section{Ablation Studies}

%-----------------------------------------------------##---------
\begin{table*}[ht]
\begin{center}
\begin{tabular}[H]{| l | c | c | c | c |}
    \hline
    Method & Price Accuracy & Title Accuracy & Image Accuracy \\
    \hline\hline
    CoVA without positional features & $ 89.2 \pm 10.3 $ & $ 91.9 \pm 1.4 $ & $ 95.9 \pm 1.8 $ \\
    \textbf{CoVA} &  $ \mathbf{95.5} \pm \mathbf{3.8} $ &  $ \mathbf{95.7} \pm \mathbf{1.2} $ & $ \mathbf{98.8} \pm \mathbf{1.5} $ \\
    \hline
\end{tabular}
\end{center}
\vspace{-0.3cm}
\caption{Importance of positional features in RN}
\label{tab:no_bbox}
\vspace{-0.3cm}
\end{table*}

\begin{table*}[hbt!]
\begin{center}
\begin{tabular}[H]{| l | c | c | c | c |}
    \hline
    Method & Price Accuracy & Title Accuracy & Image Accuracy \\
    \hline\hline
    CoVA without sampling & $ 93.0 \pm 5.6 $ &  $ \mathbf{96.0} \pm \mathbf{0.9} $ & $ 97.8 \pm 3.6 $ \\
    \textbf{CoVA} & $ \mathbf{95.5} \pm \mathbf{3.8} $ &  $ 95.7 \pm 1.2 $ & $ \mathbf{98.8} \pm \mathbf{1.5} $ \\
    \hline
\end{tabular}
\end{center}
\vspace{-0.3cm}
\caption{Improvement in performance due to sampling}
\label{tab:sampling}
\vspace{-0.3cm}
\end{table*}

\noindent\textbf{Importance of Positional features.} 
%\label{subsec:abl_pos_feat}
We train CoVA without positional features to gauge its importance. Table~\ref{tab:no_bbox} shows that positional features can significantly improve accuracy for price, title, and image prediction. This also validates that for webpage object detection, location and size of a bounding box carries significant information, making it different from classical object detection.

% \subsection{Comparison of different CNNs in RN}
% We experimented with first few layers in AlexNet \cite{krizhevsky2012imagenet} as the CNN (since it is simpler and trains much faster) for learning visual features instead of ResNet18 \cite{he2016deep} in RN. Table \ref{tab:alexnet} shows that features extracted by ResNet18 are much better than that of AlexNet.

\begin{figure}[t]
    \centering
    \includegraphics[width=0.8\linewidth]{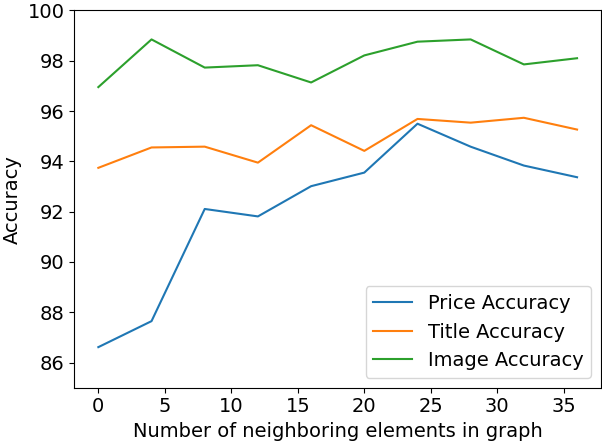}
    \caption{Comparison of context size with accuracy}
    \label{fig:context_plot}
\end{figure}

\noindent\textbf{Dependence on number of Neighbors in Graph.} 
Fig.~\ref{fig:context_plot} shows the variation in cross domain accuracy of CoVA with respect to the number of neighboring elements $ K $. Note that having $0$ context elements is equivalent to our baseline Fast R-CNN*. We observe that, unlike for title and image, 
price accuracy can significantly be improved by considering larger contexts. This is due to the fact that price is highly ambiguous (Fig.~\ref{fig:multiple_prices}). We also study the graph construction described by \cite{graph_multimodal} where all nodes are considered in the neighborhood of a particular node. This significantly reduced the performance for price ($90.7\%$) and title ($92.7\%$).

\noindent\textbf{Importance of Sampling in training.}
%\label{subsec:abl_sampling}
As discussed in Section~\ref{sec:exp}, we introduce a random sampling of $90\%$ background web elements while training.  Table~\ref{tab:sampling} shows that this leads to improvements in results.

\section{Conclusion \& Future Work}
In this paper, we reformulated the problem of webpage IE (WIE) as a context-aware webpage object detection. We created a large-scale dataset for this task, which we will release publicly.  We proposed CoVA  which uses i) a graph representation of a webpage, ii) a Representation Network (RN) to learn visual representation for a web element, and iii) a Graph Attention Network (GAT) for  contextual learning. CoVA improves upon state-of-the-art results and newly created baselines by considerable margins. Our visualizations show that CoVA is able to attend to the most important contexts.  In the future, we would like to adapt this method to other tasks such as identifying malicious web elements. 
% We would also like to explore complex approaches such as incorporating hierarchical prior from DOM tree in the Bayesian framework.

{\small
\bibliographystyle{plainnat}
\bibliography{egbib}
}

\end{document}